\documentclass[sigconf ,authordraft, nonacm,  review=False]{acmart}

\usepackage{pgfplots}
\pgfplotsset{compat=newest}

\usepackage{graphicx}
\usepackage{subcaption}
\usepackage{multirow}
\usepackage{bm}
\usepackage{tabularx}
\usepackage{amsmath}
\usepackage{array}
\usepackage{ulem}
\usepackage{xcolor}

\AtBeginDocument{%
  \providecommand\BibTeX{{%
    \normalfont B\kern-0.5em{\scshape i\kern-0.25em b}\kern-0.8em\TeX}}}

\setcopyright{acmlicensed}
\copyrightyear{2024}
\acmYear{2024}
\acmDOI{XXXXXXX.XXXXXXX}

\begin{document}

\title{Context Neural Networks: A Scalable Multivariate Model for Time Series Forecasting}

\author{ Abishek Sriramulu}

\author{Christoph Bergmeir}

\author{Slawek Smyl}

\begin{abstract}
Real-world time series often exhibit complex interdependencies that cannot be captured in isolation. Global models that model past data from multiple related time series globally while producing series-specific forecasts locally are now common. However, their forecasts for each individual series remain isolated, failing to account for the current state of its neighbouring series. Multivariate models like multivariate attention and graph neural networks can explicitly incorporate inter-series information, thus addressing the shortcomings of global models. However, these techniques exhibit quadratic complexity per timestep, limiting scalability. This paper introduces the Context Neural Network, an efficient linear complexity approach for augmenting time series models with relevant contextual insights from neighbouring time series without significant computational overhead. The proposed method enriches predictive models by providing the target series with real-time information from its neighbours, addressing the limitations of global models, yet remaining computationally tractable for large datasets. 
\end{abstract}


\maketitle

\section{Introduction}

Time series forecasting is a vital capability for many applications in finance, economics, transportation, energy usage, and other domains. Accurate forecasts enable better planning, resource allocation, and decision making. As such, developing effective and efficient time series forecasting models has become an active area of research.

Classical statistical methods like ARIMA and exponential smoothing have been used for many decades \citep{hyndman2018forecasting}. But in recent years, machine learning approaches, especially neural networks, have proven superior across many forecasting tasks \citep{hewamalage2022global}. Deep neural networks have been explored in recent works to overcome the limitations of statistical models. Recurrent Neural Network (RNN) models like LSTMs \citep{bandara2020forecasting} and GRUs, Convolutional Neural Network (CNN) based models like TCN \citep{wan2019multivariate} and LSTNet \citep{lai_modeling_2018}, Attention mechanisms \citep{Bahdanau2015NeuralMT} incorporated in RNN models like TPA-LSTM \citep{tpa_attention_Shih2019} and transformer models like Temporal Fusion Transformer \citep{lim_temporal_2019} have proven to improve modeling long-term dependencies. However, these models do not capture both inter-series and intra-series dynamics. These models are trained on large datasets comprising many related time series. They aim to learn patterns and relationships that generalize across series. However, most existing approaches utilize only the history of each individual series as input during training and inference. The core premise of this paper is that forecast accuracy can be improved by incorporating relevant contextual information from other related series. But naively adding such cross-series inputs faces computational and informational challenges.
Current best practices involve training neural network forecasting models globally, on batches of data drawn from many series \citep{hewamalage2022global}. These models learn to map historical sequences to future values, where the sequences consist of past observations or lagged values of the target series. In some cases, exogenous variables known to affect the series are also provided as inputs. But rarely are raw inputs derived from other series in the dataset directly incorporated. Intuitively, providing additional information about the concurrent state of related series could further improve performance. In our proposed approach, the global model learns general patterns that apply across all series and a Series-specific context is provided to the target series.
The main contribution of our work is the design of a scalable neural network based topology modeling for forecasting. This model serves as a more computationally efficient alternative to graph neural networks. Specifically, our context/topology modeling neural network scales more easily to large datasets while retaining strong performance in forecasting tasks. This makes it suitable for real-world systems involving complex and extensive topologies, where graph neural networks would be too computationally expensive to implement effectively. Source code of our mechanism will be made available publicly.

\section{Related Work}

\subsection{Global Models}

Global (univariate) modeling in time series forecasting means that not a single series is seen as a dataset for which a model is built, but that one shared model is built across a set of series. It has been a very successful paradigm over recent years, finally rendering complex models such as neural networks competitive for time series forecasting. 

Recurrent neural networks, such as Long Short-Term Memory (LSTM) and Gated Recurrent Unit (GRU), have emerged as one of the most popular methods in this field, achieving state-of-the-art results. They excel in processing sequential data by iteratively transforming the current hidden state and the next input into the subsequent state. An often-used paradigm in this context is the RNN encoder-decoder model, where the encoder compresses the input sequence into a fixed representation, which the decoder utilizes to generate the target sequence. For example, \citet{salinas2020deepar} introduced DeepAR, which utilizes a global RNN architecture with separate hidden states for each time series while sharing parameters across series. Having distinct hidden states for each series allows modeling individual dynamics, while parameter sharing enables exploiting common patterns. \citet{smyl2020hybrid} proposed a hybrid ES-RNN model combining exponential smoothing (ES) with RNN. ES equations are used internally to train some local per-series parameters, e.g., for seasonality, together with a subsequent global RNN.

Also Convolutional Neural Networks like Temporal Convolutional Networks (TCN) \citep{bai2018empirical} have demonstrated remarkable efficacy in univariate forecasting. TCN leverages stacked dilated causal convolutions to capture long-range dependencies in the data, with channel-wise dilations to optimize computation and residual connections to combat the vanishing gradient problem.

\subsection{Multivariate Models}

Multivariate forecasting introduces challenges beyond capturing temporal patterns, such as modeling spatial or other cross-series dependencies. Traditional methods like VAR models \citep{holden1995vector} and Gaussian Processes \citep{schulz2018tutorial} have limitations in handling complex nonlinear relationships. Recent deep learning advances offer neural network-based approaches for complex time series modeling.
Models like VectorRNN \citep{goel2017r2n2} merge VAR and RNN, utilizing separate RNNs for each series. LSTNet \citep{lai_modeling_2018} incorporates CNNs in the encoder, enhancing efficiency. MTL-RNN \citep{yuan2023covid19} and DeepDGL \citep{chen2021learning} take alternative approaches, with separate TCN encoders or shared convolutions and attention mechanisms, respectively, to exploit correlations. However, these models don't explicitly model inter-series relationships. The self-attention mechanism, popular for capturing long-range dependencies in sequence data \citep{vaswani_attention_17}, also allows explicit modeling of inter-series relationships in multivariate forecasting. However, attention has high computational complexity for large numbers of series.

Graph neural networks (GNNs) have emerged as a promising paradigm for multivariate time series forecasting. 
By connecting related time series nodes through edges, GNNs capture complex multivariate dependencies and achieve excellent empirical performance across various forecasting applications but face challenges in scalability. GNNs like Graph Convolutional Networks (GCNs) and Graph Attention Networks (GATs) like ADLGNN provide flexible relational reasoning on graph structured data \citep{wu2020connecting, sriramulu2023adaptive}. GNNs interleave graph convolution and temporal convolutions to handle both cross-series and temporal patterns.
Most GNN models rely on a predefined graph structure based on domain knowledge. Whereas in most real-word problems, a well predefined graph does not exist. To overcome this limitation, MTGNN \citep{wu2020connecting} initializes a fully-connected graph and refines edge weights through training. Models like DCRNN \citep{li2017diffusion} learn interpretable graphs from multivariate data using sequence modeling on graph structured RNNs. While most recent works focus on neural networks, researchers have also combined statistical methods with neural networks. For example, \citet{sriramulu2023adaptive} proposed a model combining  statistical structure modeling in conjunction with a deep learning approach.

\section{Problem Formulation}

Multivariate or Global RNNs and CNNs primarily rely on shared parameters and latent space projections to extract common patterns and couplings across series. Attention mechanisms directly model dependencies and GNNs provide an explicit way to encode relationships. However, both are computationally expensive.
This can be represented as learning a mapping $f$ that uses context series $r$ to minimize forecast error on target series $x$, with $t$ being the current time and $L$ a loss function:
\begin{equation}
f^* = \textit{argmin}(L\left(f(r^{in}_{(0,\ldots,t)},x^{in}_{(0,\ldots,t)}), x^{out}_{(t+1,\ldots,\tau)}\right))
\end{equation}

\subsection{Computational Challenges}

Attention mechanisms are effective for incorporating contextual information in sequence modeling tasks \citep{Bahdanau2015NeuralMT}. They have been adapted to operate across time series, allowing the model to learn which historic states in which series have the most relevance for forecasting a target series \citep{dual_self_attn_2019}. But cross-series attention is expensive, O($N^2$) per timestep, where $N$ is the number of series. This quickly becomes intractable for large $N$.

Approaches based on GNNs can constrain attention to a subset of series assumed to have strong relationships with the target \citep{wu2020connecting}. But computational costs are still high for dynamic graphs that recalculate connectivity at each timestep. Performance also depends heavily on the validity of the graph structure.

In general, any method that compares or combines information across a quadratic number of series pairs per timestep faces inherent scalability issues. This has largely restricted advanced cross-series modeling to small or moderately sized datasets in research settings.

\subsection{Informational Challenges}

Aside from computational constraints, average-based aggregation methods like graph neural networks face informational limitations. Averaging obscures nuanced relationships between series that could provide useful forecasting signals.
For example, consider three hypothetical series A, B, and C. High values in A might predict low values in C. But high values in B predict high values in C. If both A and B have increasing trend, the effects cancel out in an average, even though both series contain useful signals for forecasting C.
Some approaches address this by concatenating raw context series data, but as mentioned above, neural network inputs are order-dependent. Simply concatenating other series' histories discards vital positioning information that the model relies on to detect patterns.

\section{Model Framework}

To address these challenges, we propose ContextRNN - a type of neural network architecture that is designed to be more efficient and scalable than GNNs and Transformers, while preserving their capacity to model complex structured data.

ContextRNN handles graph or sequence data one element at a time in the RNN computational graph, rather than processing all edges simultaneously line GNNs or other multivariate models. This reduces complexity to O(N) for N series. Additionally, ContextRNN incorporates supplemental contextual information tailored to each series' relationships to avoid limitations of aggregation-based approaches. Contexts provide relevant signals from related series at each time step without averaging away nuanced dependencies. The forecast still relies primarily on the target series' own history, with contexts providing relevant signals from related series at each time step.

Furthermore, ContextRNN can natively handles smaller input windows, missing data and unequal series lengths common in real-world data. The proposed method  balances modeling expressiveness, computational performance, and practical data handling requirements for effective large-scale forecasting.

\subsection{Context Selection}
\label{Context Selection}

Context is a representation of the sequential information from related time series that are relevant to the current time step of a target series. It is analogous to the neighborhood of a target node in a GNN.
Context provides supplementary information about what is happening concurrently in other related time series at and before a given time step. The context series can be selected using a variety of methods, such as: \\
\textbf{Simple heuristics:} Based on business knowledge, similar or related time series can be identified for every target time series. For example, if we are forecasting sales for toothpaste, we could identify other personal care products, such as shampoo and mouthwash, as context series. \\
\textbf{Geography or Pre-defined structure:} Sometimes, a predefined network exists that can be used to select related time series. For example, when forecasting traffic in a city, we can select time series representing geographically close points, because these are likely to be related. \\
\textbf{Hierarchical relationships:} We can use the parent and ancestor series of a target series as context series. This can be useful for adding information from higher levels to noisy lower-level series. For example, we can aggregate all product series in a particular subcategory, category and store into context series. \\ 
\textbf{Data Driven Methods:} We can use a variety of data-driven methods to identify context series that contain information that is beneficial to the target series. These methods range from complex Bayesian network structure learning methods, such as Grow-Shrink (GS) and Incremental, Association Markov Blanket (IAMB), to relatively simpler information theory-based methods, such as Granger causality and mutual information \citep{sriramulu2023adaptive}.

 In this study, we adopt a data-driven methodology inspired by the framework proposed by \citet{sriramulu2023adaptive}, incorporating various simple information theory methods. We first obtain three adjacency matrices using the following information theory based methods: Correlation $A_\text{CM}$, Correlation Spanning Tree $A_\text{CST}$, and Mutual Information Score $A_\text{MI}$. These matrices are normalized using min-max normalization and aggregated by taking the mean, resulting in a single weighted adjacency matrix. The values in this matrix are then ranked, and the top $1.5*S$ context series for each target series are identified, where $S$ is a hyperparameter specifying the number of context series to be used in the neural network. Granger causality analysis is performed between each target series and its $1.5*S$ selected context series, reducing the number of Granger causality tests from $N^2$ (for all pairs) to $N*1.5*S$. This results in the Granger causality matrix $A_\text{GC}$. From this matrix, we select the top $S$ ranking context series for each target series. All the information theory methods were implemented using scikit-learn, an open-source Python package \citep{scikit-learn}. Additional details can be found in the supplementary material. 
The process of selecting context can be efficiently managed through non-data-driven methods such as domain knowledge or random selection, making the process operate in constant time.  In scenarios where such domain knowledge is not available or proves insufficient, we resort to data-driven methods, at which point the computational complexity escalates to quadratic. Nonetheless, this more computationally intensive step is undertaken only once as a pre-processing step before training, rendering it manageable in most real-world scenarios. 
This strategy of selecting context as a part of the pre-processing step can also be beneficial for GNNs, as it helps in simplifying the complexities involved in learning graph structures. Similar simplifications are explored by \citet{sriramulu2023adaptive}. However, GNNs still process the relationships of all series simultaneously during the training process leading to a computational complexity that scales quadratically with the number of series. Our proposed model, ContextRNN diverges from this approach. Unlike GNNs, ContextRNN does not attempt to model interactions between all the series together. Instead, it focuses on the relationship between one target series and its specific context vector at each time step. By doing so, it significantly reduces complexity, as the increase in computational demand is linear rather than quadratic.

\subsection{Neural Network Architecture}

\begin{figure*}[htb]
	\centering
	\includegraphics[width=0.7\textwidth, height=0.35\textheight]{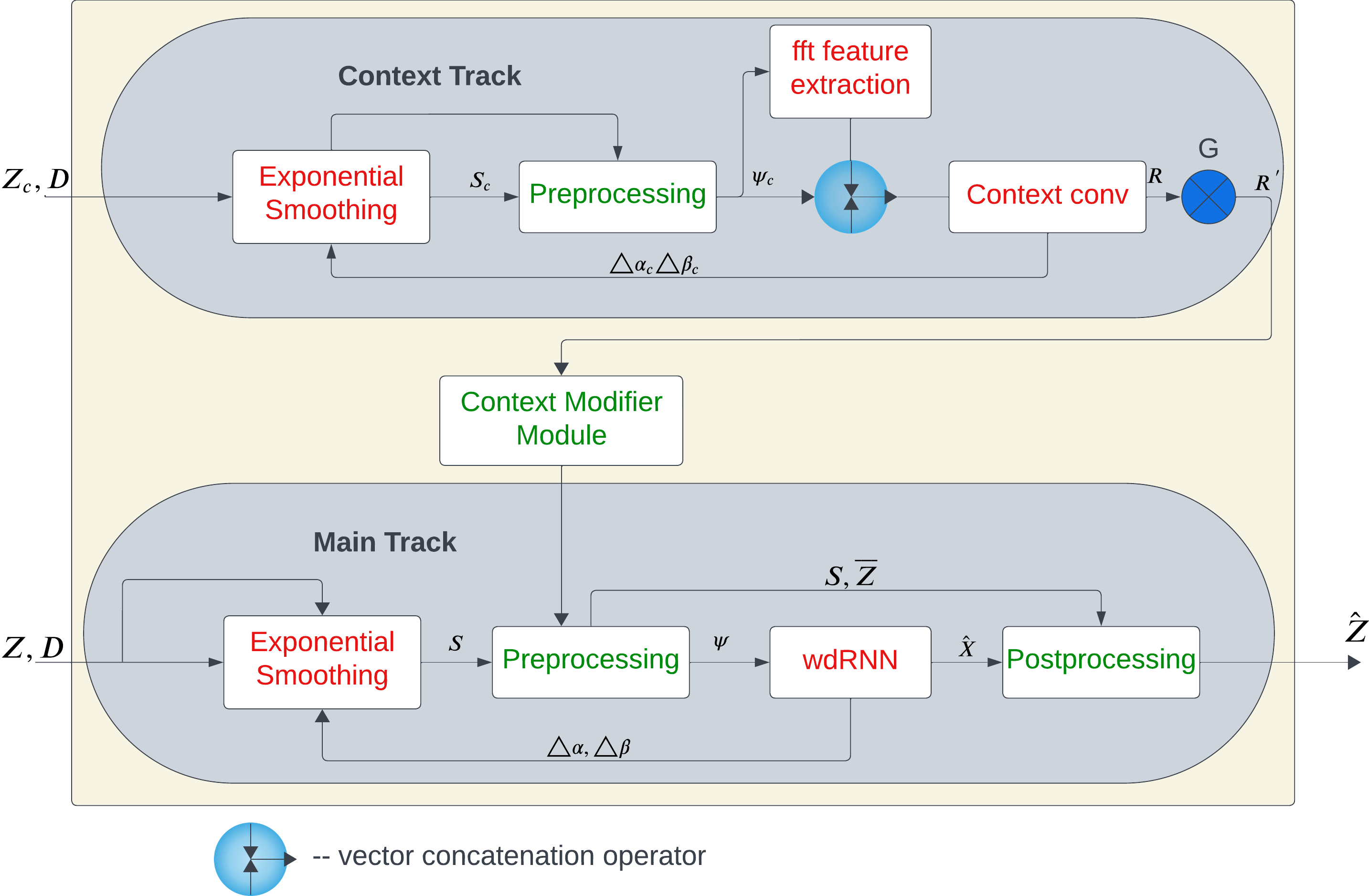}
	\caption{Model Architecture}
	\label{fig:model_arch_c3}
\end{figure*}

The proposed time series forecasting model (see Figure \ref{fig:model_arch_c3}) employs a dual-track architecture comprising a context track and a main forecasting track. This design aims to leverage both classical statistical techniques as well as modern deep learning methods for accurate modeling of complex time series.

The context track processes a representative subset of related time series to extract useful patterns that supplement the target time series. The input to this track is a batch of context series that are expected to contain patterns relevant to modeling the target time series. These could be selected based on the procedures outlined in Section~\ref{Context Selection}. This track performs preprocessing and feature extraction tasks. The context vector provides a condensed representation of historical information in the context series that can aid in forecasting the target series. The context vector is then modulated per time series to improve adaptation to individual dynamics. 

The main track is focused on modeling the target time series to generate multi-step forecasts. It takes as input the target series as well as other explanatory variables like calendar data, exogenous variables, and other inputs. Additionally, the context vector from the first track is also provided as input. The main track performs preprocessing, feature extraction and forecast generation tasks. The main track is composed of stacked layers of weighted dilated RNN cells (wdRNNCells). The unique design of wdRNNCells incorporates delayed states and an internal weighting mechanism. The delayed states extend the receptive field of the cells to capture long-range dependencies and multiple seasonalities.

The ES and preprocessing module in each track decomposes the respective time series into level and seasonal components and performs normalization to aid the modeling. The smoothing parameters are learned dynamically by the neural network. The parameters of the two tracks are tuned jointly to minimize the final forecasting error. The complete architecture comprised of the context track and main forecasting track is trained end-to-end using a pinball loss function. The model can also be trained in a cross-learning setup using time series from multiple domains or datasets. This allows the model to exploit common patterns across different series while also adapting to individual dynamics. Regularization techniques like ensembling further improve the model's generalization capability.

\subsubsection{Exponential Smoothing:} Exponential smoothing is a method that assigns exponentially decreasing weights to past observations. This means that more recent observations have a greater influence on the forecast than older observations. This module employs a simplified Holt-Winters formulation \citep{hyndman2018forecasting} to decompose the time series into trend and seasonal components. The key feature of this module is the use of dynamic smoothing parameters $\alpha_t$ and $\beta_t$ that are adjusted at each time step based on learned corrections from the neural network:
\begin{align}
\alpha_{t+1} &= \sigma(\mathbf{I}\alpha + \Delta\alpha_t) \\
\beta_{t+1} &= \sigma(\mathbf{I}\beta + \Delta\beta_t) 
\end{align}
Where $\mathbf{I}\alpha, \mathbf{I}\beta$ are the initial smoothing values, $\Delta\alpha_t, \Delta\beta_t$ are the neural network tuned corrections, and $\sigma$ is the sigmoid function. The trend and seasonal components are then updated as:
\begin{align}
l_t &= \alpha_t z_t + (1 - \alpha_t) l_{t-1} \\
s_{t+p} &= \beta_t \frac{z_t}{l_t} + (1 - \beta_t) s_t
\end{align}
Here, $z_t$ is the original series, $l_t$ is the trend, $s_t$ is the seasonal component, and $p$ is the seasonal period. Allowing the network to modulate the smoothing parameters gives a dynamically adaptive decomposition tailored to each series.

\subsubsection{Preprocessing \& Postprocessing:}

The primary input to the preprocessing module is the historical sequence immediately preceding the forecast period. Let $\Delta_{t}^{\text{in}}$  be the input window at step $t$, and $\Delta_{t}^{\text{out}}$ be the output window. The input sequence 
is preprocessed as:
\begin{equation}
x_{t,\tau}^{\text{in}} = \log\left(\frac{z_\tau}{\bar{z}_t \hat{s}_{t,\tau}}\right), \quad \tau \in \Delta_t^{\text{in}}
\end{equation}
where $\bar{z}_t$ is the average of the input sequence, and $\hat{s}_{t,\tau}$ is the seasonal component predicted by the ES module. This achieves three goals: 1) Log transform prevents outliers from skewing learning. 2) Deseasonalization with $s_{t,\tau}$ removes seasonality. 3) Normalization by $\bar{z}_t$ removes long-term trend.
Thus all preprocessed series are normalized to a similar level, which aides effective cross-learning across series. For the output, the forecast period $\Delta_t^{\text{out}}$ is simply normalized as:
\begin{equation}
x_{t,\tau}^{\text{out}} = \frac{z_\tau}{\bar{z}_t}, \quad \tau \in \Delta_t^{\text{out}}
\end{equation}
The input sequences $x_t^{\text{in}}$ and output sequences $x_t^{\text{out}}$ are generated by sliding the windows $\Delta_t^{\text{in}}$ and $\Delta_t^{\text{out}}$ forward in time. The post-processing component converts the RNN output into actual forecasts. The main forecasting track predicts a vector $\hat{x}_t^{\text{RNN}} \in \mathbb{R}^{fh}$. This is transformed into actual values $\hat{z}_\tau$ via:
\begin{equation}
\hat{z}_\tau = \exp(\hat{x}_{t,\tau}^{\text{RNN}}) \bar{z}_t \hat{s}_{t,\tau}, \quad \tau \in \Delta_t^{\text{out}}
\end{equation}
where the ES seasonal components $\hat{s}_{t,\tau}$ and input average $\bar{z}_t$ reverse the preprocessing. This postprocessing converts the normalized RNN outputs into actual forecasts. The lower and upper bounds of the predictive intervals, $\hat{x}_t^{\text{RNN}}$ and $\hat{\overline{x}}_t^{\text{RNN}}$, are converted to actual forecasts in the same fashion.

\subsubsection{Context Convolution:}
The Context Convolution module consists of a sequence of 1D convolution layers interspersed with nonlinearities and residual connections. To incorporate both time and frequency domain characteristics, the preprocessed input sequence \(x_t^{\text{in}}\) is first transformed into the frequency domain using a Fast Fourier Transform (FFT).   
\[ \hat{x}(f) = FFT\{x_t^{\text{in}}\} \]
We decompose the Fourier-transformed sequence \(\hat{x}(f)\) into its real (\(Re\)), imaginary (\(Im\)), magnitude (\(|\hat{x}(f)|\)), and phase (\(\angle \hat{x}(f)\)) components. These components are stacked together with the original preprocessed input to form a comprehensive feature set:
    \[ X_{stacked} = [Re(\hat{x}(f)), Im(\hat{x}(f)), |\hat{x}(f)|, \angle \hat{x}(f), x_t^{\text{in}}] \]
Two convolutional layers with residual connections are used to build up a representation of the context time series. The first is a depthwise convolutional layer that applies a convolution independently to each input channel. This is followed by a pointwise convolution to combine features across channels. Residual connections add the input to the output of each convolution layer before applying a nonlinearity, which allows the network to learn modifications to the identity function. After flattening the convolutional feature maps, a final linear layer reduces the dimensionality and outputs the context vector.

This module outputs a fixed-length vector $r^{'}_t$ summarizing the most relevant features from the representative time series and ES parameters $\Delta\alpha_t$, $\Delta\beta_t$. This context vector is used in the main track to provide additional global information for forecasting each individual series. The context track parameters are trained jointly with the main track to minimize the overall forecast error.

\subsubsection{Context Modifier module:}
In neural network systems, the order and position of inputs matter significantly. Simply concatenating the context with main track inputs, even if they are well chosen and likely helpful for a particular time series, does not guarantee that the network can properly utilize and integrate that information.
In our proposed model, context information is generated by a dedicated Context Convolution module processing a representative subset of time series. This produces a context vector $\mathbf{r}'_t \in \mathbb{R}^{uK}$ at each time step $t$, where $u$ is the size of the individual context vector per series and $K$ is the number of series in the context batch.
To adjust the global context $\mathbf{r}'_t$ to each individual series, we introduce per-series modulation vectors $\mathbf{g}^{(j)}_t \in \mathbb{R}^{uK}$, where $j$ indexes the series in the main forecasting batch. The modulation vectors are initialized to ones and element-wise multiply the context vector to produce a customized context for each series:
\begin{equation}
\mathbf{r}'^{(j)}_t = \mathbf{r}'_t \otimes \mathbf{g}^{(j)}_t
\end{equation}

Here, $\otimes$ denotes element-wise multiplication. The modulated context vector $\mathbf{r}'^{(j)}_t$ is then concatenated with the other per-series inputs and fed into the main forecasting RNN.
While $\mathbf{r}_t$ captures useful global patterns across the context series, the specific way that information should influence the forecast likely varies for each individual series. The modulation vectors $\mathbf{g}^{(j)}_t$ serve to adjust the context to the needs of each series $j$. Elements of $\mathbf{g}^{(j)}_t$ that are less than one attenuate the corresponding context, while elements greater than one amplify it. The modulation vectors are optimized along with all other model parameters to minimize the forecasting loss function. This allows $\mathbf{g}^{(j)}_t$ to be tuned specifically for each series $j$ based on how well it improves the forecast accuracy over many training iterations. In a sense, the model is learning how best to modulate the common context signal for each series. 
The modulated context vectors $\mathbf{r}'^{(j)}_t$ are then concatenated with the other per-series inputs $\mathbf{x}^{in}_t$ to form the input to the main forecasting RNN:
\begin{equation}
\mathbf{x}^{in'}_t = [\mathbf{x}^{in}_t, \hat{s}_t, \log_{10}(\bar{z}_t), \mathbf{d}^w_t, \mathbf{d}^m_t, \mathbf{d}^y_t, \mathbf{r}'^{(j)}_t]
\end{equation}
Here, $\mathbf{x}^{in}_t$ is the preprocessed input sequence, $\hat{s}_t$ is the seasonal component vector,
$\log_{10}(\bar{z}_t)$ is the log average value of the input sequence, $\mathbf{d}^w_t, \mathbf{d}^m_t, \mathbf{d}^y_t$ are calendar variables, $\mathbf{r}'^{(j)}_t$ is the modulated context vector.

\subsubsection{wdRNN:}
This module is designed to model the complex temporal dependencies in the time series, including short-term, long-term  and seasonality. The wdRNN module stacks multiple wdRNNCells with increasing dilations in a hierarchical structure. The increasing dilations in higher layers enable modeling dependencies at different timescales in a hierarchical manner. Short-term dependencies are captured by the first layer, while longer-term dependencies are extracted by higher layers. Residual connections are used to improve the gradient flow during training. The calendar variables in $x'_t$ are first passed through an embedding layer to project the sparse binary encoding into a dense representation $d_t$. The output layer produces the forecast $\hat{x}^{\text{RNN}}_t$, lower PI bound $\hat{x}^{\text{RNN}}_t$, upper PI bound $\hat{\bar{x}}^{\text{RNN}}_t$, and ES parameters $\Delta\alpha_t$, $\Delta\beta_t$.

The wdRNN cell used in our model refers to the adRNN cell introduced by \citet{smyl2022drnn}. It is composed of two identical dilated RNN cells (dRNNCells) stacked together. Each dRNNCell contains the following elements: \\
\textbf{Dilated states:} In addition to the standard recent state $h_{t-1}$, the cell maintains a dilated state $h_{t-d}$ from $d$ steps ago, where $d > 1$ is the dilation, to model longer-term dependencies. \\
\textbf{Dual cell states:} There are two cell states - recent $c_{t-1}$ and dilated $c_{t-d}$. The dilated state augments the cell's memory capacity. \\
\textbf{Gating mechanisms:} Similar to GRU, there are update and fusion gates to manage the flow of information between the states. The fusion gate weights the recent and dilated $c$-states, while the update gate weights the combined $c$-state and the candidate $\tilde{c}_t$. \\
\textbf{Split outputs:} The output is split into two parts - the primary output $y_t$ which is passed to the next layer, and the controlling state $h_t$ used internally in the gating equations. This provides more flexibility compared to LSTM/GRU where a single $h$-state serves both purposes. \\
\textbf{Weighting mechanism:} The bottom dRNNCell produces a weight vector $m_t$ that weights the input to the upper dRNNCell in an input-dependent manner. This enables dynamic input weighting. The first dRNNCell produces a weight vector $\mathbf{m}_t$ that weights the input $\mathbf{x}_t$:
\begin{align*}
\mathbf{f}_t &= \sigma(\mathbf{W}_f \mathbf{x}_t + \mathbf{V}_f \mathbf{h}_{t-1} + \mathbf{U}_f \mathbf{h}_{t-d} + \mathbf{b}_f) \\
\mathbf{u}_t &= \sigma(\mathbf{W}_u \mathbf{x}_t + \mathbf{V}_u \mathbf{h}_{t-1} + \mathbf{U}_u \mathbf{h}_{t-d} + \mathbf{b}_u) \\
\mathbf{o}_t &= \sigma(\mathbf{W}_o \mathbf{x}_t + \mathbf{V}_o \mathbf{h}_{t-1} + \mathbf{U}_o \mathbf{h}_{t-d} + \mathbf{b}_o) \\
\tilde{\mathbf{c}}_t &= \tanh(\mathbf{W}_c \mathbf{x}_t + \mathbf{V}_c \mathbf{h}_{t-1} + \mathbf{U}_c \mathbf{h}_{t-d} + \mathbf{b}_c) \\
\mathbf{c}_t &= \mathbf{u}_t \otimes (\mathbf{f}_t \otimes \mathbf{c}_{t-1} + (1 - \mathbf{f}_t) \otimes \mathbf{c}_{t-d}) + (1 - \mathbf{u}_t) \otimes \tilde{\mathbf{c}}_t \\
\mathbf{h}_t^{1'} &= \mathbf{o}_t^{1} \otimes \mathbf{c}_t^{1}, \quad \mathbf{m}_t = [\mathbf{h}_{t,1}^{1'}, \ldots, \mathbf{h}_{t,s_m}^{1'}], \\& \quad \mathbf{h}_t^{1} = [\mathbf{h}_{t,s_m+1}^{1'}, \ldots, \mathbf{h}_{t,s_m+s_h}^{1'}]
\end{align*}
Here, $\mathbf{f}_t$, $\mathbf{u}_t$, $\mathbf{o}_t$ are fusion, update, and output gates respectively. $\tilde{\mathbf{c}}_t$ is a candidate cell state. $\mathbf{h}_t$ and $\mathbf{m}_t$ are output states where $\mathbf{m}_t$ represents the weight vector.
The weight vector $\mathbf{m}_t$ is passed through an exponential function to produce weights $\mathbf{a}_t$:
\begin{align}
\mathbf{a}_t = \exp(\mathbf{m}_t)
\end{align}
These weights are used to modulate the input $\mathbf{x}_t$ to get the weighted input $\mathbf{x}^a_t$:
\begin{align}
\mathbf{x}^a_t = \mathbf{a}_t \otimes \mathbf{x}_t
\end{align}
The weighted input $\mathbf{x}^a_t$ is then passed to the second dRNNCell to produce the output $\mathbf{y}_t$.
By stacking two dRNNCells together with a weighting mechanism, the wdRNNCell can focus on the relevant components of the input $\mathbf{x}_t$ for each time step.

\subsection{Loss Function}

A pinball loss function is used to optimize both the point forecasts and predictive intervals (PIs) produced by the model. The loss function is defined as:
\begin{equation}
L_{\tau} = \rho(x^{\text{out}}_\tau, \hat{x}^{\text{out}}_{q^*,\tau}) + \gamma(\rho(x^{\text{out}}_\tau, \hat{x}^{\text{out}}_{\underline{q},\tau}) + \rho(x^{\text{out}}_\tau, \hat{x}^{\text{out}}_{\overline{q},\tau}))
\end{equation}
Here, $\rho$ is the pinball loss function, $x^{\text{out}}_\tau$ is the actual normalized load value, $\hat{x}^{\text{out}}_{q^*,\tau}$ is the forecasted median value, and $\hat{x}^{\text{out}}_{\underline{q},\tau}$, $\hat{x}^{\text{out}}_{\overline{q},\tau}$ are the lower and upper bounds for the predictive intervals.
The first component of the loss function corresponds to the point forecast, while the second and third components relate to the predictive interval bounds.
The hyperparameter $\gamma \geq 0$ controls the relative importance of the predictive interval components.

\section{Experimental Setup}

\subsection{Datasets and Error metrics}

We use three benchmark multivariate time series datasets that have been previously used in related works \citep{lai_modeling_2018,wu2020connecting, sriramulu2023adaptive}. This allows direct comparison of our model results to state-of-the-art and baseline methods.
We refrain from utilizing the exchange rate and solar energy production datasets also oftentimes used in these setups. It has been well established that the exchange rate use case is flawed \citep{hewamalage2023forecast}, and the solar energy dataset presents domain-specific challenges, as solar energy production drops to zero during periods of no sunlight. Addressing this challenge of many very predictable zeros typically involves the utilization of specialized pre-processing techniques, such as use of a clear sky model, which are not the main focus of the paper. \\
\textbf{RE-Europe:} Contains electricity load recorded hourly at 1494 nodes across Europe with a sequence length of 26,304 \citep{jensen2017re}.  \\ 
\textbf{Electricity:} Contains hourly electricity consumption data from 321 clients from 2012-2014 with a sequence length of 26,304 \citep{lai_modeling_2018}.  \\
\textbf{Traffic:} Contains hourly road occupancy rates from 2015-2016 across 862 freeway sensors in the San Francisco Bay area with a sequence length of 17,544 \citep{lai_modeling_2018}. \\
\textbf{PEMS-BAY:} Contains traffic flow data from 325 sensors in the San Francisco Bay area at 5-minute intervals with a sequence length of 52,116 \citep{wu2020connecting}. \\
\textbf{METR-LA:} Contains traffic flow data from 207 sensors in Los Angeles at 5-minute intervals with a sequence length of 34,272 \citep{wu2020connecting}. \\
To evaluate model performance on these datasets, we utilize two common metrics, Root Relative Squared Error (RSE) and  Correlation coefficient (CORR), as in \citet{lai_modeling_2018}.

\subsection{Baseline Methods}

The performance of the proposed model is compared with a range of baseline models, including statistical models, classical machine learning models, recurrent neural networks, deep neural networks, and graph neural networks and hybrid architectures, as follows: 
\textbf{Non-GNN Models:} AR, VAR-MLP \citep{hybrid_arima_mlp03}, GP \citep{Roberts2013GaussianPF}, PatchTST \citep{nie2022time}, RNN-GRU, LSTNet \citep{lai_modeling_2018}, TPA-LSTM \citep{tpa_attention_Shih2019}, HyDCNN \citep{li2021modeling}, SCInet \citep{liu2022scinet}, IRN \citep{choo4492823irregularity}. \\
\textbf{GNN Models:} MTGNN \citep{wu2020connecting}, ADLGNN \citep{sriramulu2023adaptive}, SDLGNN \citep{sriramulu2023adaptive}, SDLGNN-Corr, AGLG-GRU \citep{guo4439467learning}, GTS \citep{shang2021discrete}, SDGL \citep{li2023dynamic}, TDG4-MSF \citep{miao2023tdg4msf}.  \\
\textbf{ContextNN:} Our proposed model.   \\ 
More details can be found in the supplementary material.

\begin{table*}[!htb]
\caption{RSE Comparison of state-of-the-art methods. }
\label{tab:init_exp_c3}
\centering
\small
\setlength{\tabcolsep}{3pt} 

\begin{tabular}{l|cccc|cccc|cccc|cccc}

 & \multicolumn{4}{c}{\textbf{Electricity Dataset}}& \multicolumn{4}{c}{\textbf{Traffic Dataset}} & \multicolumn{4}{c}{\textbf{METR-LA}} & \multicolumn{4}{c}{\textbf{PEMS-BAY}}\\
 & \multicolumn{4}{c}{Horizons}& \multicolumn{4}{c}{Horizons} & \multicolumn{4}{c}{Horizons
} & \multicolumn{4}{c}{Horizons
}\\ 
\textbf{Methods} &  3 & 6 & 12 & 24 & 3 & 6 & 12 & 24  & 3 & 6 & 12 &24   & 3& 6 & 12 &24   \\ \hline
 \multicolumn{17}{c}{Non-Graph Neural Network models}\\ \hline
AR &  0.0995 & 0.1035 & 0.1050 & 0.1054 & 0.5911 & 0.6218
 & 0.6252 & 0.6300  & 0.7943& 0.7460& 0.8751&0.9502 & 0.6924& 0.6889& 0.7492&0.7704
\\ 
VAR-MLP &  0.1393 & 0.162 & 0.1557 & 0.1274 & 0.5582 & 0.6579 & 0.6023 & 0.6146  & 0.7399& 0.7641& 0.8736&0.9610 & 0.6718& 0.7112& 0.7383&0.7872
\\ 
GP &  0.1500 & 0.1907 & 0.1621 & 0.1273 & 0.6082 & 0.6772 & 0.6406 & 0.5995  & 0.6915& 0.7516& 0.9144&0.9396 & 0.7243& 0.7278 & 0.7415&0.7970 
\\ 
PatchTST & 0.2407 & 0.1910 & 0.2978 & 0.1166 & 0.6965 & 0.6780 & 0.7772 & 0.5911  & 0.7584& 0.8046& 0.8165&0.8584 & 0.6497  & 0.7149 & 0.7694&0.7773
\\
RNN-GRU &  0.1102 & 0.1144 & 0.1183 & 0.1295 & 0.5358 & 0.5522 & 0.5562 & 0.5633  & 0.7155& 0.7081& 0.7581&0.7398 & 0.6254 & 0.6304& 0.6519&0.6576 
\\ 
LSTNet &  0.0864 & 0.0931 & 0.1007 & 0.1007 & 0.4777 & 0.4893 & 0.495 & 0.4973  & 0.6549& 0.6494& 0.6716&0.6961 & 0.5661 & 0.5698& 0.5830&0.5961
\\ 
TPA-LSTM &  0.0823 & 0.0916 & 0.0964 & 0.1006 & 0.4487 & 0.4658 & 0.4641 & 0.4765  & 0.6146& 0.6530& 0.6598&0.6640 & 0.5320 & 0.5587& 0.5620&0.5708
\\ 

HyDCNN &  0.0832 & 0.0898 & 0.0921 & 0.094 & 0.4198 & 0.429 & 0.4352 & 0.4423  & 0.5687& 0.5944& 0.6010&0.6114 & 0.4940 & 0.5114& 0.5186&0.5265
\\ 

SCInet & 0.0748 & 0.0845 & 0.0926 & 0.0976 & 0.4216 & 0.4414 & 0.4495 & 0.4453  & 0.5909& 0.5981& 0.6154&0.6340 & 0.5078 & 0.5195& 0.5326&0.5398
\\ 
IRN & 0.0739 & 0.0844 & 0.0926 & 0.0968 & 0.4171 & 0.4349 & 0.4493 & 0.4449  & 0.5977& 0.6007& 0.6318&0.6665 & 0.5081 & 0.5180& 0.5407&0.5563
\\ \hline
 \multicolumn{17}{c}{Graph Neural Network Models}\\ \hline

MTGNN &  0.0745 & 0.0878 & 0.0916 & 0.0953 & 0.4162 & 0.4754 & 0.4461 & 0.4535  & 0.6130& 0.6302& 0.6556&0.6588 & 0.5159 & 0.5222& 0.5417&0.5611
\\ 
SDLGNN &  \textit{0.0726} & 0.0820 & 0.0896 & 0.0947 & 0.4053 & 0.4209 & 0.4313 & 0.4444  & 0.5549& 0.5628& 0.5786&0.6034 & 0.4804 & 0.4913& 0.5043 &0.5247
\\ 

SDLGNN-CORR& 0.0737 & 0.0841 & 0.0923 & 0.0971 & 0.4227 & 0.4378 & 0.4576 & 0.4579  & 0.5601& 0.5799& 0.5970&0.6075 & 0.4911 & 0.5087& 0.5276&0.5320
\\ 

AGLG-GRU &  0.0738 & 0.0864 & 0.0912 & 0.0947 & 0.4173 & 0.4722 & 0.4427 & 0.4526  & 0.6113& 0.6129& 0.6576&0.6608 & 0.5158 & 0.5428& 0.5502&0.5560
\\ 

GTS &  0.0790 & 0.0884 & 0.0957 & 0.0951 & 0.4665 & 0.4779 & 0.4792 & 0.4766  & 0.5923& 0.6209& 0.6374&0.6433 & 0.5298 & 0.5499& 0.5587&0.5596
\\ 

SDGL &  \textbf{0.0698} & \textbf{0.0805} & \textit{0.0889} & \textit{0.0935} & 0.4142 & 0.4475 & 0.4584 & 0.4571  & 0.5627& 0.5719& 0.5970&0.6100 & 0.4881 & 0.5099& 0.5278&0.5338
\\ 
TDG4-MSF & 0.0731 & 0.0828  & 0.0894 & 0.0969 &\textbf{0.4029} & \underline{0.4196} & \underline{0.4294} & \underline{0.4366}  & \textit{0.5483}& \textit{0.5597}& \underline{0.5731}&\underline{0.6020}& \underline{0.4669}& \underline{0.4803}& \textit{0.5044}&\textit{0.5214}\\

ADLGNN & \underline{0.0719} & \underline{0.0809} & \underline{0.0887} & \underline{0.0930} & \underline{0.4047} & \textit{0.4201} & \textit{0.4299} &\textit{0.4416}  & \underline{0.5284}& \underline{0.5412}& \textit{0.5783}&\textit{0.6021}& \textit{0.4753}& \textit{0.4892}& \underline{0.5015}&\underline{0.5188}\\ \hline
 \multicolumn{17}{c}{Our Proposed Model}\\ \hline
ContextRNN & 0.0738 & \textit{0.0813} & \textbf{0.0868} & \textbf{0.0914} & \textit{0.4051} & \textbf{0.4182} & \textbf{0.4292} & \textbf{0.4340}  &\textbf{ 0.5132}& \textbf{0.5265}&\textbf{ 0.5451}&\textbf{0.5679} & \textbf{0.4493}& \textbf{0.4628}& \textbf{0.4769}&\textbf{0.4903}\\ 
\end{tabular}

\end{table*}

\begin{table*}
\caption{Evaluating Model Performance and Scalability on High-Dimensional Large-Scale Dataset}
    \centering
\small
\setlength{\tabcolsep}{3pt} 

    \begin{tabular}{c|ccccc|r}
         \textbf{Methods}&  \multicolumn{4}{c}{\textbf{RSE/CORR per Horizon}} & \textbf{Mean} & \textbf{Minutes}\\
         &   3&  6&  12& 24 & \textbf{RSE} & \textbf{per Epoch}\\ \hline

			MTGNN &  0.0501/0.9544& 0.0613/0.9366& 0.0742/0.9237& 0.0692/0.9346&0.0637 &33.38\\

            SDLGNN &  0.0459/0.9792& 0.0574/0.9612& 0.0677/0.9366& 0.0687/0.9352&0.0599 &20.01\\

            SDLGNN-CORR & 0.0467/0.9786& 0.0511/0.9617& 0.0696/0.9257& 0.0671/0.9365&0.0586 &19.23\\

            AGLG-GRU &  0.0486/0.9784& 0.0564/0.9563& 0.0687/0.9347& 0.0693/0.9258&0.0608 &30.65\\

            GTS &  0.0543/0.9612& 0.0658/0.9366& 0.0749/0.9195& 0.0762/0.9137&0.0678 &28.11\\

            SDGL &  0.0483/0.9795& 0.0578/0.9590& 0.0675/0.9362& 0.0686/0.9353&0.0605 &30.26\\

            TDG4-MSF &  0.0388/0.9819& 0.0503/0.9669& 0.0549/0.9601& 0.0598/0.9597&0.0509 &13.52\\

			ADLGNN &  0.0386/0.9850& 0.0499/0.9754& 0.0641/0.9623& 0.0681/0.9571&0.0552 &24.78\\

         ContextRNN&  \textbf{0.0229/0.9894}&  \textbf{0.0312/0.9873}&  \textbf{0.0413/0.9809}& \textbf{0.0507/0.9645}&\textbf{0.0365} &\textbf{6.51}\\

    \end{tabular}
    
    \label{tab:add_exp}
 
\end{table*}

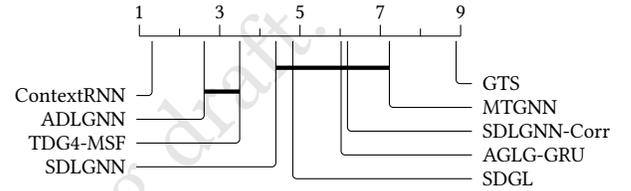
\begin{figure}[htb]
	\centering

\begin{tikzpicture}[
  treatment line/.style={rounded corners=1.5pt, line cap=round, shorten >=1pt},
  treatment label/.style={font=\small},
  group line/.style={ultra thick},
]

\begin{axis}[
  clip={false},
  axis x line={center},
  axis y line={none},
  axis line style={-},
  xmin={1},
  ymax={0},
  scale only axis={true},
  width={0.24\textwidth},
  ticklabel style={anchor=south, yshift=1.3*\pgfkeysvalueof{/pgfplots/major tick length}, font=\small},
  every tick/.style={draw=black},
  major tick style={yshift=.5*\pgfkeysvalueof{/pgfplots/major tick length}},
  minor tick style={yshift=.5*\pgfkeysvalueof{/pgfplots/minor tick length}},
  title style={yshift=\baselineskip},
  xmax={9},
  ymin={-5.5},
  height={4.5\baselineskip},
  xtick={1,3,5,7,9},
  minor x tick num={1},
]

\draw[treatment line] ([yshift=-2pt] axis cs:1.3165455544566654, 0) |- (axis cs:0.8676470588235294, -2.5)
  node[treatment label, anchor=east] {ContextRNN};
\draw[treatment line] ([yshift=-2pt] axis cs:2.61248027580542097, 0) |- (axis cs:0.8676470588235294, -3.5)
  node[treatment label, anchor=east] {ADLGNN};
\draw[treatment line] ([yshift=-2pt] axis cs:3.49847258520584651, 0) |- (axis cs:0.8676470588235294, -4.5)
  node[treatment label, anchor=east] {TDG4-MSF};
\draw[treatment line] ([yshift=-2pt] axis cs:4.39822054774805325, 0) |- (axis cs:0.8676470588235294, -5.5)
  node[treatment label, anchor=east] {SDLGNN};
\draw[treatment line] ([yshift=-2pt] axis cs:4.82048578835240765, 0) |- (axis cs:9.338235294117647, -6.0)
  node[treatment label, anchor=west] {SDGL};
\draw[treatment line] ([yshift=-2pt] axis cs:6.0240974855235086, 0) |- (axis cs:9.338235294117647, -5.0)
  node[treatment label, anchor=west] {AGLG-GRU};
\draw[treatment line] ([yshift=-2pt] axis cs:6.18245887530946702, 0) |- (axis cs:9.338235294117647, -4.0)
  node[treatment label, anchor=west] {SDLGNN-Corr};
\draw[treatment line] ([yshift=-2pt] axis cs:7.22584755386900427, 0) |- (axis cs:9.338235294117647, -3.0)
  node[treatment label, anchor=west] {MTGNN};
\draw[treatment line] ([yshift=-2pt] axis cs:8.88824505876342059, 0) |- (axis cs:9.338235294117647, -2.0)
  node[treatment label, anchor=west] {GTS};
\draw[group line] (axis cs:2.61248027580542097, -2.3333333333333335) -- (axis cs:3.49847258520584651, -2.3333333333333335);
\draw[group line] (axis cs:4.39822054774805325, -1.3333333333333333) -- (axis cs:7.22584755386900427, -1.3333333333333333);

\end{axis}
\end{tikzpicture}
	\caption{The critical difference diagram implemented following \citet{demvsar2006statistical} comparing the mean ranks of all GNN models and ContextRNN.}
	\label{fig:cdDia}
\end{figure}

\subsection{Evaluation Setup}

The baseline model outcomes except for PatchTST on the Electricity and Traffic datasets were acquired from previously published works, specifically \citep{lai_modeling_2018,sriramulu2023adaptive,miao2023tdg4msf,li2023dynamic,guo4439467learning,shang2021discrete,li2021modeling,liu2022scinet,choo4492823irregularity}. For PatchTST, experiments were conducted using default hyperparameters. For the RE-Europe, PEMS-BAY and METR-LA datasets all the baseline experiments were conducted using hyperparameters listed on their respective literatures . In all experimental configurations, the most recent 20\% of the data was designated as the test set, the initial 60\% served as the training set, and the remaining 20\% was allocated for validation purposes. Details on hyperparameters and training setup of our proposed model can be found in the supplementary material.

\section{Results}

Table \ref{tab:init_exp_c3} summarises the RSE on the test data for all the methods on electricity and traffic datasets. To highlight the top three best performing methods, the table uses boldface font for the method with the lowest (best) RSE, underlining for the second lowest RSE, and italics for the third lowest RSE. The proposed model demonstrates superior performance compared to current state-of-the-art models across various forecast horizons (For CORR, see Table \ref{tab:init_exp_c3_cor} in the supplementary material). Particularly noteworthy is its advantage for long forecast horizons, such as 24 hours, where it achieves a reduction in RSE by 3.4\% compared to state-of-the-art methods. Furthermore, despite its enhanced forecasting accuracy, the proposed model retains relatively low complexity, ensuring efficiency and practicality in implementation.

To further evaluate our method against GNN approaches, additional experiments were conducted using the RE-Europe dataset to investigate the performance on a larger dataset. The experiments to obtain the RSE metric were run on multiple machines with different hardware configurations, so the results cannot be used for runtime comparison. Hence, to acquire the Minutes/Epoch metric, dedicated experiments were conducted on a machine with 16 vCPUs, 64GiB RAM, 1TiB SSD, and 1 × NVIDIA GeForce RTX 4090 (24GB) GPU. All GNN methods were run for 3 epochs with a batch size of 4, which was the largest size that could be accommodated by the 24GB GPU memory for the RE-Europe dataset. Since ContextRNN is lightweight, large batch size was not a concern, so 100 was used.

The results from the additional experiments presented in Table \ref{tab:add_exp} on both RSE and CORR metrics align with the findings from table~\ref{tab:init_exp_c3}. Our proposed model outperforms current state-of-the-art methods. Specifically in the RE-Europe dataset, our model reduces the RSE by 28.29\% compared to the best existing GNN model. This demonstrates that our approach is more effective for large datasets. Additionally, our model is much more efficient, needing only 48.15\% of the computation time required by the state-of-the-art technique. This corresponds to a 51.85\% reduction in runtime compared to TDG5-MSF and 73.73\% compared to ADLGNN. This significant decrease in computational demand makes our model particularly scalable and suitable for large graph data.

The critical difference diagram presented in Figure~\ref{fig:cdDia} provides a visual representation of the statistical analysis conducted to evaluate the statistical significance of differences in performance between the proposed ContextRNN model and a suite of state-of-the-art GNN models. This diagram distinctly positions ContextRNN as the leading model, exhibiting statistically significant outperformance over its GNN counterparts, providing robust evidence of its efficacy. The diagram shows ContextRNN with the lowest average rank, indicating superior performance relative to all other models. Notably, models such as ADLGNN, and TDG4-MSF are grouped closely, suggesting no significant differences among them, while ContextRNN stands apart without overlap, underscoring its statistical dominance. The methodology for this statistical comparison is grounded in the framework established by \citet{demvsar2006statistical}. The strategic decision to exclude non-GNN methods from the statistical analysis stems from the fact that the error metrics for non-GNN methods on electricity and traffic datasets were acquired from existing literature and did not have access to the detailed forecast data from their studies. Furthermore, their exclusion was deemed justified on the grounds that, as evidenced by the outcomes presented in Table~\ref{tab:init_exp_c3}, these models did not demonstrate competitive performance, failing to rank within the top five across any dataset or forecast horizon analyzed. The results of the critical difference diagram reinforce this decision by clearly demonstrating that ContextRNN significantly outperforms the GNN models.

\section{Ablation study}

\begin{figure}
    \centering
    \includegraphics[width=1.1\linewidth]{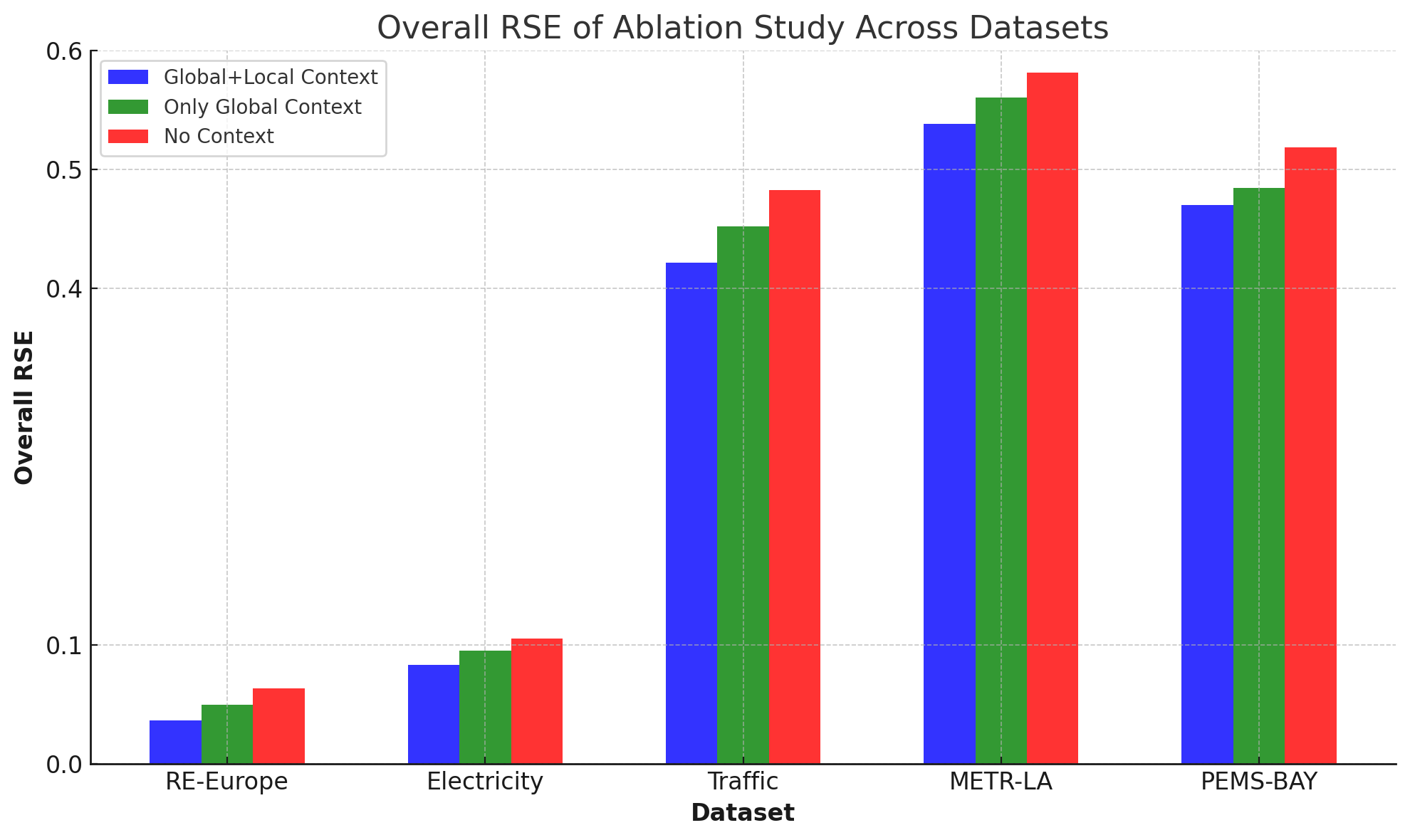}
    \caption{Ablation Study}
    \label{fig:Ablation Study}
\end{figure}

The ablation study across multiple datasets, including RE-Europe, Electricity, Traffic, METR-LA, and PEMS-BAY, reveals significant insights into the efficacy of incorporating contextual information within forecasting models. As illustrated in Figure~\ref{fig:Ablation Study}, we examine three distinct configurations: model with both global and local context (context convolution + context modifier), model with only global context (context convolution but no context modifier), and model without any context (no context convolution \& no context modifier). In Figure~\ref{fig:Ablation Study}, the relative improvements observed from the addition of both global and local contexts underscore a substantial enhancement in forecasting accuracy. Specifically, the transition from only a global context to a combined global and local context results in a decrease in RSE by 26.46\% for the RE-Europe dataset, 12.32\% for the Electricity dataset, 6.73\% for the Traffic dataset, 3.91\% for the METR-LA dataset, and 2.99\% for the PEMS-BAY dataset. This enhancement is even more pronounced when comparing the global and local context configuration against model with no context, with relative decreases in RSE by 42.50\% for RE-Europe dataset, 20.82\% for Electricity dataset, 12.64\% for Traffic dataset, 7.40\% for METR-LA dataset, and 9.41\% for PEMS-BAY dataset.

\section{Conclusion}

We have presented a novel neural network architecture for multivariate time series forecasting that is designed to capture both cross-series and temporal dependencies in an efficient manner, while also being robust to missing data and variable series lengths. Through extensive experiments on real-world datasets, we have demonstrated that our approach outperforms existing methods, in particular with significant improvement on a large dataset. The primary breakthrough in our model lies in its lightweight design, allowing for robust predictive performance without the substantial computational overhead of graph neural networks. Our work represents an important advancement towards scalable, accurate forecasting for multivariate time series across a variety of domains.

\begin{acks}
This research was supported by the Australian Research Council under grant DE190100045, a Facebook Statistics for Improving Insights and Decisions research award, Monash University Graduate Research funding, and the MASSIVE - High performance computing facility, Australia. Christoph Bergmeir is currently supported by a María Zambrano (Senior) Fellowship that is funded by the Spanish Ministry of Universities and Next Generation funds from the European Union.
\end{acks}
\bibliographystyle{ACM-Reference-Format}
\bibliography{mybibfile}

\appendix
\section{Appendix: Supplementary Material}
\subsection{Context Selection Methods - Additional details}
\textbf{Correlation:} Correlation is simple and fast but measures only linear relationships.
We use the Pearson correlation coefficient to measure linear relationships between each pair of series.

\textbf{Correlation Spanning Tree (CST):} The graph is constructed by building a minimum spanning tree connecting all the series in the data using the distance matrix computed from Pearson's correlation matrix \citep{mantegna1999hierarchical}. The correlation spanning tree converts the distance matrix of correlations into a minimum spanning tree connecting all series. First, the distance matrix is computed from correlations:
\begin{equation}
D_{ij} = 1 - A_{CM_{ij}}
\end{equation}
Single linkage hierarchical clustering is applied, as follows. 1) Start with each series in its own cluster  2) Iteratively merge the closest clusters $C_1$, $C_2$ based on the minimum distance $\min_{x\in C_1, y\in C_2} D_{xy}$.
This builds a dendrogram of merged clusters. Cutting at height 1 produces the minimum spanning tree with edges between merged clusters. Higher correlations result in shorter distances and tree edges.

\textbf{Mutual Information (MI):} The adjacency matrix is obtained by estimating the mutual information $I(X;Y)$ between the probability distributions of each pair of series in the data. Mutual information measures by how much knowledge of one variable reduces uncertainty about the other:
\begin{equation}
I(X;Y) = \int \int p(x,y) \text{log}\left(\frac{p(x,y)}{p(x)p(y)}\right) dx dy
\end{equation}

\textbf{Granger Causality (GC):} Granger causality analysis is a statistical technique used to determine whether one time series helps predict or 'Granger causes' another time series. This method can identify context series that improve forecasting accuracy for a target series. The Granger causality procedure involves building multiple augmented models, each including a different candidate context series along with the target series. These models are applied to a validation dataset. Models with context series that outperform the baseline model without contexts are retained. F-tests are then conducted to evaluate the statistical significance of the predictive improvement gained by including each context series. Contexts are ranked based on the $p$-values from the F-tests, with lower $p$-values indicating stronger Granger causality relationships. This process systematically determines which contexts demonstrate significant Granger causality and are most useful for forecasting the target series.

\subsection{Baseline Model Descriptions}

\begin{description}
    \item[AR:] Auto-regressive linear model.
	\item[VAR-MLP:] A composite model of a linear VAR and an MLP \citep{hybrid_arima_mlp03}. 
	\item[GP:] A Gaussian Process Model \citep{Roberts2013GaussianPF}. 
    \item[PatchTST]: Transformer-based multivariate time series forecasting model \citep{nie2022time}
	\item[RNN-GRU:] RNN with GRU.
	\item[LSTNet:] A neural network model with temporal convolution layers in recurrent fashion with skip connections \citep{lai_modeling_2018}.
	\item[TPA-LSTM:] RNN with temporal attention \citep{tpa_attention_Shih2019}.
	\item[HyDCNN:] This model uses a position-aware fully dilated CNN to model the non-linear patterns, followed by an autoregressive model which captures the sequential linear dependencies \citep{li2021modeling}.
\item[SCInet:] SCINet is a recursive downsample-convolve-interact neural network that uses multiple convolutional filters to extract temporal features from the downsampled sub-sequences \citep{liu2022scinet}.
 \item[IRN:] Irregularity Reflection Neural Network is a CNN based forecasting model that captures unexpected changes in time series data by interpreting them as irregularities using a Fourier series approach \citep{choo4492823irregularity}.
 
 \item[MTGNN:] A GNN model that first constructs the graph from data and uses the learned inter-series dependencies to make forecasts better \citep{wu2020connecting}.
 \item[ADLGNN:] A GNN model that uses information theory based methods to initialize the graph and uses an attention mechanism to keep the graph dynamic over time \citep{sriramulu2023adaptive}.
 \item[SDLGNN:] A static variant of ADLGNN (excludes attention mechanism).
\item[SDLGNN-Corr:] Static variant of ADLGNN, which uses a pairwise Pearson correlation as graph.

\item[AGLG-GRU:] Similar to ADLGNN, Adaptive Global-Local Graph Structure Learning with Gated Recurrent Units \citep{guo4439467learning}.
  
 \item[GTS:] A Recurrent GNN model that learns a probabilistic graph through optimizing the mean performance over the graph distribution. The distribution is parameterized by a neural network so that discrete graphs can be sampled differentiably through reparameterization \citep{shang2021discrete}.
 \item[SDGL:] A GNN model that uses a dynamic graph learning method to generate time-varying matrices based on changing node features and static node embedding \citep{li2023dynamic}.
\item[TDG4-MSF:] A GNN forecasting model that uses temporal decomposition enhanced representation learning \citep{miao2023tdg4msf}.

\end{description}

\subsection{Hyperparameters \& training set-up}

The model training procedure employed for all experiments using the proposed model was as follows: 

The model was trained for 11 epochs, with batch size progressively increased from 2 to 5 at epoch 4, from 5 to 12 at epoch 5, from 12 to 25 at epoch 6, from 25 to 50 at epoch 7, and finally from 50 to 100 at epoch 8. A step-wise decreasing learning rate schedule was used, beginning at $3\times10^{-3}$ for epochs 1-8, decreasing to $10^{-3}$ at epoch 9 and $10^{-4}$ for epochs 10-11. The Adam optimizer was utilized for training. Training was initialized from random starting points in order to increase variation. The model employed dilated RNN cells with dilations of [2, 6, 12, 24], suitable for the hourly resolution of the datasets used for experimentation. The datetime encoder size was set to 74 with an embedding size of 8. The context vector size $u$ was 2. The number of context series $K$ was set to 15 for all datasets except RE-Europe, for which $K=5$ due to the limited availability of useful relationships between time series. The PI weight $\gamma$ was set to 0.4. Since the experimentation compares the results with point forecasting models only, a single quantile was used. In particular, we use the quantile 0.48. We use a value different from 0.5 to counter a systematic bias empirically found in the experiments, stemming from the logarithm that is applied as a preprocessing to all series. Input window size was set to be 168.

\subsection{Additional outcomes of experiments}

\begin{table*}[!htb]
\caption{CORR Comparison of state-of-the-art methods. }
\label{tab:init_exp_c3_cor}
\centering
\small
\setlength{\tabcolsep}{3pt} 
\begin{tabular}{c|cccc|cccc|c}

 & \multicolumn{4}{c}{\textbf{Electricity Dataset}}& \multicolumn{4}{c}{\textbf{Traffic Dataset}}& \textbf{Overall Mean} \\
 & \multicolumn{4}{c}{Horizons}& \multicolumn{4}{c}{Horizons}&\\ 
\textbf{Methods} &  3 & 6 & 12 & 24 & 3 & 6 & 12 & 24 & \\ \hline
AR & 0.8845 &	0.8632 & 0.8591 & 0.8595 & 0.7752 &	0.7568 & 0.7544 & 0.7519 & 0.8131 \\ 
VAR-MLP & 0.8708 &	0.8389 & 0.8192 & 0.8679 &	0.8245& 0.7695 & 0.7929 & 0.7891 & 0.8216 \\ 
GP & 0.8670&0.8334&0.8394&0.8818&0.7831&0.7406&0.7671&0.7909&0.8130 \\ 
PatchTST &0.8013&0.8298&0.7981&0.8647&0.7492&0.7418&0.7183&0.8288&0.7915 \\
RNN-GRU & 0.8597&0.8623&0.8472&0.8651&0.8511&0.8405&0.8345&0.8300&0.8488 \\ 
LSTNet &  0.9283&0.9135&0.9077&0.9119&0.8721&0.8690&0.8614&0.8588&0.8903 \\ 
TPA-LSTM &0.9439&0.9337&0.9250&0.9133&0.8812&0.8717&0.8717&0.8629&0.9004 \\ 
MTGNN &0.9474&0.9316&0.9278&0.9234&0.8960&0.8667&0.8794&0.8810&0.9066\\ 
HyDCNN &0.9354&0.9329&0.9285&0.9264&0.8915&0.8855&0.8858&0.8819&0.9085 \\ 
AGLG-GRU &0.9434&0.9302&0.9283&0.9274&0.8958&0.8541&0.8755&0.8842&0.9049\\ 
GTS &  0.9291&0.9187&0.9135&0.9098&0.8685&0.8582&0.8589&0.8573&0.8893 \\ 
SDGL & 0.9534&0.9445&0.9351&0.9301&0.9010&0.8825&0.8760&0.8766&0.9124 \\ 
SCInet &0.9492&0.9386&0.9304&0.9274&0.8920&0.8809&0.8772&0.8825&0.9098 \\ 
IRN & 0.9493&0.9390&0.9313&0.9281&0.8920&0.8861&0.8774&0.8788&0.9102 \\ 
TDG4-MSF &0.9499&0.8371&0.9306&0.9246&0.9014&0.8925&0.8864&0.8834&0.9007 \\ 
SDLGNN-CORR&0.9475&0.9346&0.9263&0.9227&0.8937&0.8846&0.8746&0.8784&0.9078 \\ 
SDLGNN &0.9502&0.9384&0.9304&0.9257&0.9017&0.8925&0.8868&0.8801&0.9132\\ 
ADLGNN &0.9506&0.9386&0.9312&0.9294&0.9028&0.8928&0.8876&0.8818&0.9143\\ 
ContextRNN &0.9498&0.9384&0.9315&0.9298&0.9029&0.8944&0.8881&0.8838&0.9148\\ 
\end{tabular}

\end{table*}

\end{document}